\documentclass{llncs}
\usepackage{url}
\usepackage{mathtools}
\usepackage{graphicx}
\usepackage{multirow}
\usepackage{color} 
\usepackage{amsfonts}
\usepackage[draft]{todonotes}   
\usepackage{hyperref}

\definecolor{red}{rgb}{.7,0,0}
\definecolor{green}{rgb}{0,.65,0}
\begin{document}

\title{Legal Question Answering using Ranking SVM and Deep Convolutional Neural Network}

\author{Phong-Khac Do\inst{1} \and Huy-Tien Nguyen\inst{1}
Chien-Xuan Tran\inst{1} \and \\Minh-Tien Nguyen\inst{1,2} \and Minh-Le Nguyen\inst{1}}

\institute{School of Information Science,\\Japan Advanced Institute of Science and Technology (JAIST), \\1-1 Asahidai, Nomi, Ishikawa, 923-1292, Japan.\\
\and Hung Yen University of Education and Technology (UTEHY), Hung Yen, Vietnam. \\
\email{\{phongdk, ntienhuy, chien-tran,  tiennm, nguyenml\}@jaist.ac.jp}}

\maketitle              

\begin{abstract}
This paper presents a study of employing Ranking SVM and Convolutional Neural Network for two missions: legal information retrieval and question answering in the Competition on Legal Information Extraction/Entailment. For the first task, our proposed model used a triple of features (LSI, Manhattan, Jaccard), and is based on paragraph level instead of article level as in previous studies. In fact, each single-paragraph article corresponds to a particular paragraph in a huge multiple-paragraph article. For the legal question answering task, additional statistical features from information retrieval task integrated into Convolutional Neural Network contribute to higher accuracy.



\end{abstract}
\section{Introduction}
Legal text, along with other natural language text data, e.g. scientific literature, news articles or social media, has seen an exponential growth on the Internet and in specialized systems. Unlike other textual data, legal texts contain strict logical connections of law-specific words, phrases, issues, concepts and factors between sentences or various articles. Those are for helping people to make a correct argumentation and avoid ambiguity when using them in a particular case. Unfortunately, this also makes information retrieval and question answering on legal domain become more complicated than others.

There are two primary approaches to information retrieval (IR) in the legal domain \cite{maxwell et al}: manual knowledge engineering (KE) and natural language processing (NLP). In the KE approach, an effort is put into translating the way legal experts remember and classify cases into data structures and algorithms, which will be used for information retrieval. Although this approach often yields a good result, it is hard to be applied in practice because of time and financial cost when building the knowledge base. In contrast, NLP-based IR systems are more practical as they are designed to quickly process terabytes of data by utilizing NLP techniques. However, several challenges are presented when designing such system. For example, factors and concepts in legal language are applied in a different way from common usage \cite{dan et al}. Hence, in order to effectively answer a legal question, it must compare the semantic connections between the question and sentences in relevant articles found in advance \cite{kim et al.}.

Given a legal question, retrieving relevant legal articles and deciding whether the content of a relevant article can be used to answer the question are two vital steps in building a legal question answering system. Kim et al. \cite{kim et al.} exploited Ranking SVM with a set of features for legal IR and Convolutional Neural Network (CNN) \cite{kim yoon et al} combining with linguistic features for question answering (QA) task. However, generating linguistic features is a non-trivial task in the legal domain.
Carvalho et al. \cite{dan et al} utilized n-gram features to rank articles by using an extension of TF-IDF. For QA task, the authors adopted AdaBoost \cite{Freund 1997} with a set of similarity features between a query and an article pair \cite{Nguyen et al2015} to classify a query-article pair into ``YES" or ``NO". However, overfitting in training may be a limitation of this method.
Sushimita et al. \cite{sushmita et al} used the voting of Hiemstra, BM25 and PL2F for IR task. Meanwhile, Tran et al. \cite{vutran et al} used Hidden Markov model (HMM) as a generative query model for legal IR task. 
Kano \cite{kano} addressed legal IR task by using a keyword-based method in which the score of each keyword was computed from a query and its relevant articles using inverse frequency. After calculating, relevant articles were retrieved based on three ranked scores. These methods, however, lack the analysis of feature contribution, which can reveal the relation between legal and NLP domain. This paper makes the following contributions:
\begin{itemize}
	\item We conduct detailed experiments over a set of features to show the contribution of individual features and feature groups. Our experiments benefit legal domain in selecting appropriate features for building a ranking model.
    \item We analyze the provided training data and conclude that: (i) splitting legal articles into multiple single-paragraph articles, (ii) carefully initializing parameters for CNN significantly improved the performance of legal QA system, and (iii) integrating additional features in IR task into QA task leads to better results.
    \item We propose to classify a query-article pair into ``YES'' or ``NO'' by voting, in which the score of a pair is generated from legal IR and legal QA model.
\end{itemize}
In the following sections, we first show our idea along with data analysis in the context of COLIEE. Next, we describe our method for legal IR and legal QA tasks. After building a legal QA system, we show experimental results along with discussion and analysis. We finish by drawing some important conclusions.

\section{Proposed Method}
\subsection{Basic Idea}
\begin{figure}[h!]
  	\centering
  	\includegraphics[width=\textwidth]{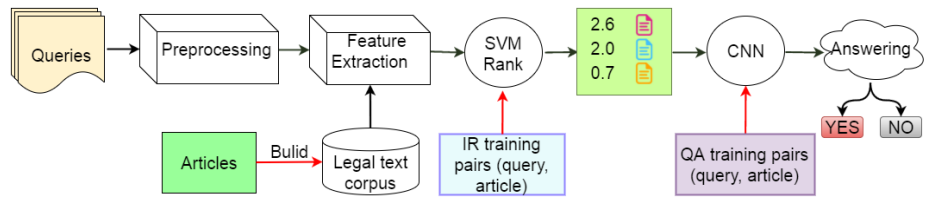}
   	\caption{The proposed model overview}
    \label{fig:coliee_proposed_model}\vspace{-0.8cm}
\end{figure}

In the context of COLIEE 2016, our approach is to build a pipeline framework which addresses two important tasks: IR and QA.
In Figure \ref{fig:coliee_proposed_model}, in training phase, a legal text corpus was built based on all articles. Each training query-article pair for LIR task and LQA task was represented as a feature vector. Those feature vectors were utilized to train a learning-to-rank (L2R) model (Ranking SVM) for IR and a classifier (CNN) for QA. The red arrows mean that those steps were prepared in advance. In the testing phase, given a query $q$, the system extracts its features and computes the relevance score corresponding to each article by using the L2R model. Higher score yielded by SVM-Rank means the article is more relevant. As shown in Figure \ref{fig:coliee_proposed_model}, the article ranked first with the highest score, i.e. 2.6, followed by other lower score articles. After retrieving a set of relevant articles, CNN model was employed to determine the ``YES" or ``NO" answer of the query based on these relevant articles.\vspace{-0.2cm}

\subsection{Data Observation}\label{sec:data}
The published training dataset in COLIEE 2016\footnote{http://webdocs.cs.ualberta.ca/$\sim$miyoung2/COLIEE2016/} consists of a text file containing Japanese Civil Code and eight XML files. Each XML file contains multiple pairs of queries and their relevant articles, and each pair has a label ``YES" or ``NO", which confirms the query corresponding to the relevant articles. There is a total of 412 pairs in eight XML files and 1,105 articles in the Japanese Civil Code file, and each query can have more than one relevant articles.

After analyzing the dataset in the Civil Code file, we observed that the content of a query is often more or less related to only a paragraph of an article instead of the entire content. Based on that, each article was treated as one of two types: single-paragraph or multiple-paragraph, in which a multiple-paragraph article is an article which consists of more than one paragraphs. There are 7 empty articles, 682 single-paragraph articles and the rest are multiple-paragraph.

Based on our findings, we proposed to split each multiple-paragraph article into several independent articles according to their paragraphs. For instance, in Table \ref{tab:multiple-article}, the Article 233 consisting of two paragraphs was split into two single-paragraph articles 233(1) and 233(2). After splitting, there are in total 1,663 single-paragraph articles. 
\begin{table}	
	\centering
	\caption{\label{tab:multiple-article} Splitting a multiple-paragraph article into some single-paragraph articles}
    \begin{tabular}{c p{10.3cm}}
        \hline
			\textbf{Article ID} & \multicolumn{1}{c}{\textbf{Content}} \\
  		\hline
        233 & (1) If a tree or bamboo branch from neighboring land crosses a boundary line, the landowner may have the owner of that tree or bamboo sever that branch. 
        (2) If a tree or bamboo root from neighboring land crosses a boundary line, the owner of the land may sever that root. \\
        \hline
        233(1) & If a tree or bamboo branch from neighboring land crosses a boundary line, the landowner may have the owner of that tree or bamboo sever that branch.\\
        \hline
        233(2) & If a tree or bamboo root from neighboring land crosses a boundary line, the owner of the land may sever that root. \\
        \hline
    \end{tabular}\vspace{-0.3cm}
\end{table}

Stopwords were also removed before building the corpus. Text was processed in the following order: tokenization, POS tagging, lemmatization, and stopword removal\footnote{http://www.nltk.org/book/ch02.html}. 
In \cite{dan et al}, the \textit{stopword removal} stage was done before the \textit{lemmatization} stage, but we found that after lemmatizing, some words might become stopwords, for instance, ``done" becomes ``do". Therefore, the extracted features based on words are more prone to be distorted, leading to lower ranking performance if \textit{stopword removal} is carried out before \textit{lemmatization} step. Terms were tokenized and lemmatized using NLTK\footnote{http://www.nltk.org/}, and POS tagged by Stanford Tagger\footnote{http://nlp.stanford.edu/software/tagger.shtml}. 

\subsection{Legal Question Answering}
\subsection*{Legal Information Retrieval}
In order to build a legal IR, traditional models such as TF-IDF, BM25 or PL2F can be used to generate basic features for matching documents with a query. Nevertheless, to improve not only the accuracy but also the robustness of ranking function, it is essential to take into account a combination of fundamental features and other potential features. Hence, the idea is to build a L2R model, which incorporates various features to generate an optimal ranking function.

Among different L2R methods, Ranking SVM (SVM-Rank) \cite{joachim}, a state-of-the-art pairwise ranking method and also a strong method for IR \cite{Zhe Cao et al,Nguyen et al2016}, was used. Our model is an extended version of Kim's model \cite{kim et al.}  with two new aspects. Firstly, there is a big distinction between our features and Kim's features. While Kim used three types of features: lexical words, dependency pairs, and TF-IDF score; we conducted a series of experiments to discover a set of best features among six features as shown in Table \ref{tab: features}. Secondly, our model is applied to individual paragraphs as described in section \ref{sec:data} instead of the whole articles as in Kim's work. 

Given n training queries $\{q_i\}_{i=1}^{n}$, their associated document pairs $(x_u^{(i)},x_v^{(i)})$ and the corresponding ground truth label $y_{u,v}^{(i)}$, SVM Rank optimizes the objective function shown in Equation \eqref{eq:objective} subject to constraints \eqref{eq:contraint-1}, and \eqref{eq:constraint-2}: 
\begin{equation}\label{eq:objective}
	min \quad \frac{1}{2}\|w\|^2 + \lambda \sum_{i=1}^{n}\sum_{u,v:y_{u,v}^{(i)}} \xi_{u,v}^{(i)}
\end{equation}
\begin{equation}\label{eq:contraint-1}
	s.t. \quad w^T(x_u^{(i)} - x_v^{(i)}) \geq 1 - \xi_{u,v}^{(i)}  \quad \text{if} \quad y_{u,v}^{(i)}=1 
\end{equation}
\begin{equation}\label{eq:constraint-2}
	\xi_{u,v}^{(i)}\geq 0, \quad i=1,...,n
\end{equation}
where: $f(x)=w^Tx$ is a linear scoring function, $(x_u,x_v)$ is a pairwise and $\xi_{u,v}^{(i)}$ is the loss. The document pairwise in our model is a pair of a query and an article.

Based on the corpus constructed from all of the single-paragraph articles (see Section \ref{sec:data}), three basic models were built: TF-IDF, LSI and Latent Dirichlet Allocation (LDA) \cite{blei et all}.  Note that, LSI and LDA model transform articles and queries from their TF-IDF-weighted space into a latent space of a lower dimension. For COLIEE 2016 corpora, the dimension of both LSI and LDA is 300 instead of over 2,100 of TF-IDF model. Those features were extracted by using \texttt{gensim} library \cite{radim et al}. Additionally, to capture the similarity between a query and an article, we investigated other potential features described in Table \ref{tab: features}. Normally, the Jaccard coefficient measures similarity between two finite sets based on the ratio between the size of the intersection and the size of the union of those sets. However, in this paper, we calculated Generalized Jaccard similarity as: 
\begin{equation}	
	J(q,A) = J(X,Y) = \frac{\sum_{i}^{} min(x_i,y_i)}{\sum_{i}^{} max(x_i,y_i)} 
\end{equation}
and Jaccard distance as:
\begin{equation}	
	D(q,A) = 1 - J(q,A)
\end{equation}
where $X = \{x_1,x_2,..,x_n\}$ and $Y=\{y_1,y_2,...,y_n\}$ are two TF-IDF vectors of a query $q$ and an article $A$ respectively.\vspace{-0.4cm}
\begin{table}
	\centering
    \setlength{\tabcolsep}{8pt}
	\caption{\label{tab: features} Similarity features for Ranking SVM}
    \begin{tabular}{l l}
        \hline
        \multicolumn{1}{l}{\textbf{Feature}} & \multicolumn{1}{c}{\textbf{Description}} \\
        \hline
        TF-IDF &  Term frequency and inverse document frequency\\
        Euclidean & Euclidean distance computed from TF \\ 
        Manhattan & Manhattan distance computed from TF \\
        Jaccard & Jaccard distance computed from TF-IDF \\
        LSI &  Latent Semantic Indexing computed from TF or TF-IDF\\
        LDA &  Latent Dirichlet Allocation computed from TF\\
        \hline
    \end{tabular}\vspace{-0.4cm}
\end{table}

The observation in Section \ref{sec:data} also indicates that one of the important properties of legal documents is the reference or citation among articles. In other words, an article could refer to the whole other articles or to their paragraphs. In \cite{dan et al}, if an article has a reference to other articles, the authors expanded it with words of referential ones. In our experiment, however, we found that this approach makes the system confused to rank articles and leads to worse performance. Because of that, we ignored the reference and only took into account individual articles themselves. The results of splitting and non-splitting are shown in Table \ref{tab:coliee 2016}.

\subsection*{Legal Question Answering}
Legal Question Answering is a form of textual entailment problem \cite{Dagan et al}, which can be viewed as a binary classification task. To capture the relation between a question and an article, a set of features can be used. In the COLLIE 2015, Kim \cite{kim yoon et al} efficiently applied Convolution Neural Network (CNN) for the legal QA task. However, the small dataset is a limit of deep learning models. Therefores, we provided additional features to the CNN model. 

The idea behind the QA is that we use CNN \cite{kim et al.} with additional features. This is because: (i) CNN is capable to capture local relationship between neighboring words, which helps CNN to achieve excellent performance in NLP problems \cite{yih et al,kim et al.,shen et al,kalchbrenner et al} and (ii) we can integrate our knowledge in legal domain in the form of statistical features, e.g. TF-IDF and LSI.
\begin{figure}[h!]
  	\centering
  	\includegraphics[width=\textwidth]{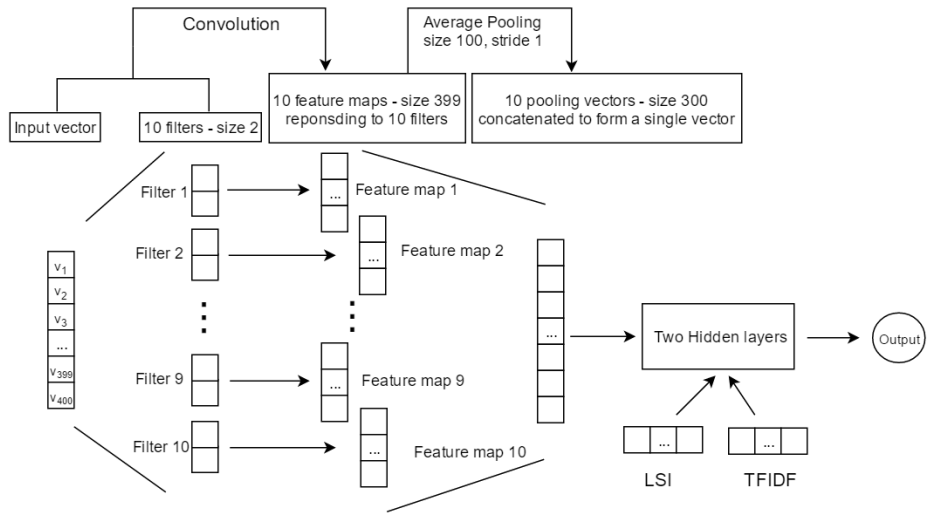}\vspace{-0.1cm}
   	\caption{The illustration of CNN model with additional features: LSI and TF-IDF. Given an input vector, CNN applies 10 filers (length = 2) to generate 10 feature maps (length = 399). Afterward, an average pooling filter (length = 100) is employed to produce average values from 4 feature maps. Finally, the average values with LSI and TF-IDF are used as input of two hidden neural network layers for QA }
    \label{fig:cnn-tfidf-lsi} \vspace{-0.3cm}
\end{figure}




In Figure \ref{fig:cnn-tfidf-lsi}, the input features $v_1,v_2,...,v_{400}$ are constructed and fed to the network as follows :
 \begin{itemize}
 \setlength{\itemindent}{2em}
   \item $v_1,v_3,v_5,...,v_{399}$: a word embedding vector of the question sentence
  \item $v_2,v_4,...,v_{400}$: a word embedding vector of the most relevant article sentence
\end{itemize}
A sentence represented by a set of words was converted to a word embedding vector $v_1^{200}$ by using bag-of-words model (BOW) \cite{yu et al}. BOW model generates a vector representation for a sentence by taking a summation over embedding of words in the sentence. The vector is then normalized by the length of the sentence:
\begin{equation}
	s= \frac{1}{n}\sum_{i= 1}^{n}s_{i}
\end{equation}
where: $s$ is a $d$-dimensional vector of a sentence, $s_{i}$ is a $d$-dimensional vector of $i^{th}$ word in the sentence, $n$ is the length of sentence. A word embedding model ($d=200$) was trained by using Word2Vec\cite{Mikolov 2013} on the data of Japanese law corpus\cite{dan et al}. The corpus contains all Civil law articles of Japan's constitution\footnote{www.japaneselawtranslation.go.jp} with 13.5 million words from 642 cleaned and tokenized articles.

A filter was denoted as a weight vector $w$ with length $h$; $w$ will have $h$ parameters to be estimated. For each input vector $S \in \mathbb{R}^{d} $, the feature map vector $O \in \mathbb{R}^{d-h+1}$ of the convolution operator with a filter $w$ was obtained by applying repeatedly $w$ to sub-vectors of $S$:
\begin{equation}
	o_{i}=w\cdot S[i:i+h-1]
\end{equation}
where: $i=0,1,2,...,d-h+1$ and ($\cdot$) is dot product operation.

Each feature map was fed to a pooling layer to generate potential features by using the average mechanism \cite{boureau et al}. These features were concatenated to a single vector for classification by using Multi-Layer Perceptron with sigmoid activation. During training process, parameters of filters and perceptrons are learned to optimize the objective function.



In our model, 10 convolution filters (length = 2) were applied to two adjacent input nodes because these nodes are the same feature type. An average pooling layer (length = 100) is then utilized to synthesize important features. 
To enhance the performance of CNN, two additional statistic features: TF-IDF and LSI were concatenated with the result of the pooling layer, then fed them into a 2-layer Perceptron model to predict the answer.

\section{Results and Discussion}

\subsection{Information Retrieval}
\subsubsection*{Training model:}
For information retrieval task, 20\% of query-article pairs are used for evaluating our model while the rest is for training. As we only consider single-paragraph articles in the training phase, if a multiple-paragraph article is relevant, all of its generated single-paragraph articles will be marked as relevant. In addition, the label for each query-article pair is set either 1 (relevant) or 0 (irrelevant). In our experiment, instead of selecting top $k$ retrieved articles as relevant articles, we consider a retrieved article $A_i$ as a relevant article if its score $S_i$ satisfies Equation \eqref{eq:s_i}: 
\begin{equation}\label{eq:s_i}
	\frac{S_i}{S_0} \geq 0.85
\end{equation}
where: $S_0$ is the highest relevant score. In other words, the score ratio of a relevant article and the most relevant article should not be lower than 85\% (choosing the value 0.85 for this threshold is simply heuristic based). This is to prevent a relevant article to have a very low score as opposed to the most relevant article. \vspace{-0.3cm}
\begin{table}[t]
\centering
\setlength{\tabcolsep}{8pt}
\caption{\label{tab: feature selection} F1-score with different feature groups with the best parameters of SVM-Rank}
\label{my-label}
\begin{tabular}{l p{4.3cm} c}
\hline
\multicolumn{1}{c}{\textbf{Method}} & \multicolumn{1}{c}{\textbf{Features}} & \textbf{F1-score}            \\ 
\hline
\multirow{2}{*}{Cosine similarity} & TF-IDF   & 0.5217  \\  
		& LSI                                    & 0.4456   \\ 
\hline
\multirow{4}{*}{SVM-Rank}       & TF-IDF, Manhattan, Jaccard                & 0.582 $\pm$ 0.018            \\
                                   & TF-IDF, Euclidean, Jaccard                   & 0.587 $\pm$ 0.011            \\ 
                                   & LDA, Manhattan, Jaccard                  & 0.598 $\pm$ 0.011            \\ 
                                   & LSI, Manhattan, Jaccard                 & \textbf{0.603 $\pm$ 0.005} \\ 
\hline
\end{tabular}\vspace{-0.3cm}
\end{table}

We ran SVM-Rank with different combinations of features listed in Table \ref{tab: features}, but due to limited space, we only report the result of those combinations which achieved highest F1-score. We compared our method to two baseline models TF-IDF and LSI which only use Cosine similarity to retrieve the relevant articles. Results from Table \ref{tab: feature selection} indicate that \textit{(LSI, Manhattan, Jaccard)} is the triple of features which achieves the best result and the most stability.


\subsubsection*{Feature Contribution:}
The contribution of each feature was investigated by using leave-one-out test. Table \ref{tab: feature contribution} shows that when all six features are utilized, the F1-score is approximately 0.55. However when excluding Jaccard, F1-score drops to around 0.5. In contrast, when other features are excluded individually from the feature set, the result remains stable or goes up slightly. From this result, we conclude that Jaccard feature significantly contributes to SVM-Rank performance.\vspace{-0.3cm}

\begin{table}[ht]
	\centering
    \setlength{\tabcolsep}{6pt}
    \caption{\label{tab: feature contribution} F1-score when excluding some features from all feature set}
    \begin{tabular}{l c | l c}
		\hline
        \multicolumn{1}{c}{\textbf{Feature}} & \textbf{F1-score} & \multicolumn{1}{c}{\textbf{Feature Groups}} & \textbf{F1-score}\\
        \hline
        All & 0.5543 & All & 0.5543\\
        All \textbackslash \{TF-IDF\} & 0.5761 & All \textbackslash \{TF-IDF, Manhattan, Jaccard\} & 0.3220\\
        All \textbackslash \{Euclidean\} & 0.5638 & All \textbackslash \{TF-IDF, Euclidean, Jaccard\} & 0.4891\\
        All \textbackslash \{Manhattan\} & 0.5543 & All \textbackslash \{LDA, Manhattan, Jaccard\} & 0.4207\\
        All \textbackslash \{Jaccard\} & \textbf{0.5069} & All \textbackslash \{LSI, Manhattan, Jaccard\} & 0.4506\\
        All \textbackslash \{LDA\} & 0.5652 & \multicolumn{1}{c}{---} & \multicolumn{1}{c}{---}\\
        All \textbackslash \{LSI\} & 0.5652 & \multicolumn{1}{c}{---} & \multicolumn{1}{c}{---}\\ \hline
	\end{tabular}\vspace{-0.3cm}
\end{table}

We also analyzed the contribution of feature groups to the performance of SVM-Rank. When removing different triples of features from the feature set, it can be seen that \textit{(TF-IDF, Manhattan, Jaccard)} combination witnesses the highest loss. Nevertheless, as shown in Table \ref{tab: feature selection}, the result of \textit{(LSI, Manhattan, Jaccard)} combination is more stable and better. \vspace{-0.1cm}

\subsubsection*{Splitting vs. Non-Splitting:}
As mentioned, we proposed to split a multiple-paragraph article into several single-paragraph articles. Table \ref{tab:coliee 2016} shows that after splitting, the F1-score performance increases by 0.05 and 0.04 with references and without references respectively. In both cases (with and without the reference), using single-paragraph articles always results a higher performance. \vspace{-0.3cm}
\begin{table}
	\centering
    \setlength{\tabcolsep}{8pt}
	\caption{\label{tab:coliee 2016} IR results with various methods in COLIEE 2016.}
    \begin{tabular} {l c}
        \hline
        \multicolumn{1}{c}{\textbf{Method}} & \multicolumn{1}{c}{\textbf{F1-score}}\\ \hline
        Non-splitting + With references & 0.5326\\
        Non-splitting + No references & 0.5652\\
        With splitting + With references & 0.5870\\
        With splitting + No references & \textbf{0.6087}\\
        \hline
    \end{tabular}\vspace{-0.3cm}
\end{table}

Results from Table \ref{tab:coliee 2016} also indicate that expanding the reference of an article negatively affects the performance of our model, reducing the F1-score by more than 0.02. This is because if we only expand the content of an article with the content of referential one, it is more likely to be noisy and distorted, leading to lower performance. Therefore, we conclude that a simple expansion of articles via their references does not always positively contribute to the performance of the model. 
\subsubsection*{Tuning Hyperparmeter:}
Since \emph{linear kernel} was used to train the SVM-Rank model, the role of trade-off training parameter was analyzed by tuning $C$ value from 100 to 2000 with step size 100. Empirically, F1-score peaks at 0.6087 with $C$ = 600 when it comes to COLIEE 2016 training dataset. We, therefore, use this value for training the L2R model.

\subsection*{Formal run phase 1 - COLIEE 2016}
In COLIEE 2016 competition, Table 6 shows the top three systems and the baseline for the formal run in phase 1 \cite{Kim Mi et al. 2016}. Among 7 submissions, iLis7 \cite{Kim et al. 2016} was ranked first with outstanding performance (0.6261) by exploiting ensemble methods for legal IR. Several features such as syntactic  similarity, lexical similarity, semantic similarity, were used as features for two ensemble methods Least Square Method (LSM) and Linear Discriminant Analysis (LDA). \vspace{-0.3cm}
\begin{table}[ht]
	\centering
    \setlength{\tabcolsep}{8pt}
	\caption{\label{tab:formal_run_2016_phase1} Formal run in phase 1, COLIEE 2016.}
    \begin{tabular}{l c}
        \hline
        \multicolumn{1}{c}{\textbf{Method}} & \multicolumn{1}{c}{\textbf{F1-score}}\\
        \hline
        TF-IDF with lemmarization (baseline) & 0.5310 \\
        HUKB-2 \cite{Onodera et al. 2016}& 0.5532\\
        iLis7 \cite{Kim et al. 2016}& 	\textbf{0.6261} \\
        Our model (JNLP3) & 0.5478 \\ \hline
    \end{tabular}\vspace{-0.3cm}
\end{table}

HUKB-2 \cite{Onodera et al. 2016} used a fundamental feature BM25 and applied mutatis mutandis for articles. If both an article and a query have conditional parts, they are divided into two parts like conditional parts and the rest part before measuring their similarity. This investigation in conditional parts is valuable since it is a common structure in laws. Their F1-score in formal rune is the second highest (0.5532), which is slightly higher than our system (0.5478) using SVM-Rank and a set of features \emph{LSI, Manhattan, Jaccard}. This shows that for phase 1, our model with a set of defined features is relatively competitive.


\subsection{Legal Question Answering}
\subsubsection*{Compared Results:}
In Legal QA task, the proposed model was compared to the original CNN model and separate TF-IDF, LSI features. For evaluation, we took out 10\% samples from training set for validation, and carried out experiments on dataset with balanced label distribution for training set, validation set and testing set. \vspace{-0.3cm}
\begin{table}
	\centering
    \setlength{\tabcolsep}{8pt}
	\caption{\label{tab:result_phase2}Phase 2 results with different models}
    \begin{tabular} {l c}
    	\hline
        \multicolumn{1}{c}{\textbf{Method}} & \multicolumn{1}{c}{\textbf{Accuracy\%}}\\\hline
        Convolutional neural network & 51.5\\
        Convolutional neural network with TF-IDF & 53.0\\
        Convolutional neural network with LSI & 54.5\\
        Our model & \textbf{57.6}\\ \hline
    \end{tabular}\vspace{-0.3cm}
\end{table}

In CNN models, we found that these models are sensitive to the initial value of parameters. Different values lead to large difference in results ($\pm$ 5\%). Therefore, each model was run $n$ times (n=10) and we chose the best-optimized parameters against the validation set. Table \ref{tab:result_phase2} shows that CNN with additional features performs better. Also, CNN with LSI produces a better result as opposed to CNN with TF-IDF. We suspect that this is because TF-IDF vector is large but quite sparse (most values are zero), therefore it increases the number of parameters in CNN and consequently makes the model to be overfitted easily.

\subsubsection*{Tuning Hyperparmeter:}
To achieve the best configuration of CNN architecture, the original CNN model was run with different settings of  number filter and hidden layer dimension. According to Table \ref{tab:cnn_tuning}, the change of hyperparameter does not significantly affect to the performance of CNN. We, therefore, chose the configuration with the best performance and least number of parameters: 10 filters and 200 hidden layer size.\vspace{-0.3cm}
\begin{table}
	\centering
    \setlength{\tabcolsep}{8pt}
	\caption{\label{tab:cnn_tuning}Results of the original CNN with different settings}
    \begin{tabular} {c c c c}
    	\hline
        \textbf{Hidden Layer Size} & \textbf{Filter 5}& \textbf{Filter 10} & \textbf{Filter 15} \\ \hline
        \multicolumn{1}{c}{100} & \multicolumn{1}{c}{51.00}& \multicolumn{1}{c}{ 51.00}& \multicolumn{1}{c}{ 51.00 }\\
        \multicolumn{1}{c}{150} & \multicolumn{1}{c}{51.00}& \multicolumn{1}{c}{ 51.00}& \multicolumn{1}{c}{ 51.00 }\\
		\multicolumn{1}{c}{200} & \multicolumn{1}{c}{51.00}& \multicolumn{1}{c}{ 51.51}& \multicolumn{1}{c}{ 51.51 }\\
        \multicolumn{1}{c}{250} & \multicolumn{1}{c}{51.00}& \multicolumn{1}{c}{ 51.51}& \multicolumn{1}{c}{ 51.51 }\\	\hline

    \end{tabular}\vspace{-0.7cm}
\end{table}

\subsection{Legal Question Answering System}
In this stage, we illustrate our framework on COLIEE 2016 data. The framework was trained on XML files, from H18 to H23 and tested on XML file H24. Given a legal question, the framework first retrieves top five relevant articles and then transfers the question and relevant articles to CNN classifier. The running of framework was evaluated with 3 scenarios:
\begin{itemize}
  \item \textit{No voting}: taking only a top relevant article to use for predicting an answer for that question.
  \item \textit{Voting without ratio}: each of results, which is generated by applying our Textual entailment model to each article, gives one vote to the answer which it belongs to. The final result is the answer with more votes.
  \item \textit{Voting with ratio}: similar to \textit{Voting without ratio}. However, each of results gives one vote corresponding to article's relevant score. The final result is the answer with higher voting score.\vspace{-0.6cm}
\end{itemize}
\begin{table}
	\centering
	\caption{\label{tab:Task 3}Task 3 results with various scenarios}
    \begin{tabular} {l c}
    	\hline
    	\multicolumn{1}{c}{\textbf{Scenario}} & \multicolumn{1}{c}{\textbf{Accuracy\%}}\\
        \hline
        No voting & 45.6\\
        Voting without ratio & \textbf{49.4}\\
        Voting with ratio & 48.1\\ \hline
    \end{tabular}\vspace{-0.3cm}
\end{table}
Table \ref{tab:Task 3} shows results with different scenarios. The result of \textit{No voting} approach is influenced by IR task's performance, so the accuracy is not as high as using voting. The relevant score disparity between the first and second relevant article is large, which causes a worse result of \textit{Voting with ratio} compared to \textit{Voting without ratio}.

\subsection*{Formal run phase 2 \& 3 - COLIEE 2016}
Table \ref{tab:formal_run_2016_phase23} lists the state-of-the art methods for the formal run 2016 in phase 2 and 3. In phase 2, two best systems are iList7 and KIS-1. iList7 applies major voting of decision tree, SVM and CNN with various features; KIS-1 just uses simple rules of subjective cases and an end-of-sentence expression. In phase 3, UofA achives the best score. It extracts the article segment which related to the query. This system also performs paraphrasing and detects condition-conclusion-exceptions for the query/article. From the experimental results, deep learning models do not show their advantages in case of a small dataset. On the other hand, providing handcraft features and rules are shown to be useful in this case. \vspace{-0.3cm}
\begin{table}[h!]
	\centering
    \setlength{\tabcolsep}{8pt}
	\caption{\label{tab:formal_run_2016_phase23} Formal run in phase 2 \& 3, COLIEE 2016.}
    \begin{tabular}{l c c}
        \hline
        \multicolumn{1}{c}{\textbf{Method}} & \multicolumn{1}{c}{\textbf{Phase 2}} & \multicolumn{1}{c}{\textbf{Phase 3}}\\
        \hline        
        iLis7 \cite{Kim et al. 2016} & \textbf{0.6286}&0.5368\\
        KIS-1 \cite{Ryosuke et al. 2016} & \textbf{0.6286}&0.5158 \\
        UofA\cite{Mi-Young Kim UofA. 2016} & 0.5571 & \textbf{0.5579}\\
        Our model (JNLP3) & 0.4857 & 0.4737\\ \hline
    \end{tabular}\vspace{-0.7cm}
\end{table}

\section{Splitting and non-splitting error analysis}
In this section, we show an example in which our proposed model using single-paragraph articles gives a correct answer in contrast with utilizing non-splitting one. Given a query with id H20-26-3: ``\textit{A mandate contract is gratuitous contract in principle, but if there is a special provision, the mandatary may demand renumeration from the mandator.}'', which refers to Article 648: \vspace{-0.4cm}
\begin{table}
    \centering
    \begin{tabular}{c p{10.3cm}}
        \hline
        \textbf{Article} & \multicolumn{1}{c}{\textbf{Content}} \\ \hline				
        648(1) & In the absence of any special agreements, the mandatary may not claim 								remuneration from the mandator. \\
        \hline
        648(2) & In cases where the mandatary is to receive remuneration, the mandatary may not claim the same 			until and unless he/she has performed the mandated business; provided, however, that if the 				remuneration is specified with reference to period, the provisions of Paragraph 2 of Article 624 shall 			apply mutatis mutandis. \\ 
        \hline
        648(3) & If the mandate terminates during performance due to reasons not attributable to the mandatary, 		the mandatary may demand remuneration in proportion to the performance already completed. \\
        \hline
    \end{tabular}
\end{table} \vspace{-0.6cm}
\begin{table}[h!]
    \centering
    \caption{\label{tab: exp_splitting}An example of retrieval articles between two methods: Splitting and Non-splitting}
    \begin{tabular}{c|c|c}
        \textbf{Method} & \textbf{Article} & \textbf{Rank} \\ \hline
        \multirow{4}{*}{Splitting} & 648(1) & 1 \\ \cline{2-3}
                                    & 653 & 2 \\ \cline{2-3}
                                    & 648(3) & 5 \\ \cline{2-3}
                                    & 648(2) & 29 \\ \hline
        					
        \multirow{2}{*}{Non-splitting} & 653 & 1 \\ \cline{2-3}
        & 648 & 6 \\ \hline
    \end{tabular}
\end{table}\vspace{-0.4cm}

Apparently, three paragraphs and the query share several words namely \textit{mandatary}, \textit{remuneration}, etc. In this case, however, the correct answer is only located in paragraph 1, which is ranked first in the single-paragraph model in contrast to two remaining paragraphs with lower ranks, $5^{th}$ and $29^{th}$ as shown in Table \ref{tab: exp_splitting}. \vspace{-0.3cm}
\begin{table}[h!]	
    \centering
    \begin{tabular}{c p{10.3cm}}
        \hline
        \textbf{Article} & \multicolumn{1}{c}{\textbf{Content}} \\ \hline				
         \multirow{6}{*}{653} & A mandate shall terminate when \\
			& (i) The mandator or mandatary dies; \\
			& (ii) The mandator or mandatary is subject to a ruling for the commencement of bankruptcy 		procedures; \\
			& (iii) The mandatary is subject to an order for the commencement of guardianship. \\
        \hline
    \end{tabular}
\end{table} \vspace{-0.3cm}

Interestingly, Article 653 has the highest relevant score in non-splitting method and rank $2^{nd}$ in splitting approach. The reason for this is that Article 653 shares other words like \textit{mandatary}, \textit{mandator} as well. Therefore, it makes retrieval system confuse and yield incorrect order rank. By using single-paragraph, the system can find more accurately which part of the multiple-paragraph article is associated with the query's content.
\section{Conclusion}
This work investigates Ranking SVM model and CNN for building a legal question answering system for Japan Civil Code. Experimental results show that feature selection affects significantly to the performance of SVM-Rank, in which a set of features consisting of \textit{(LSI, Manhattan, Jaccard)} gives promising results for information retrieval task. For question answering task, the CNN model is sensitive to initial values of parameters and exerts higher accuracy when adding auxiliary features. 

In our current work, we have not yet fully explored the characteristics of legal texts in order to utilize these features for building legal QA system. Properties such as references between articles or structured relations in legal sentences should be investigated more deeply. In addition, there should be more evaluation of SVM-Rank and other L2R methods to observe how they perform on this legal data using the same feature set. These are left as our future work.  

\section*{Acknowledgement}

This work was supported by JSPS KAKENHI Grant number 15K16048, JSPS KAKENHI Grant Number JP15K12094, and CREST, JST.

\end{document}